# Using NLP to build the hypertextuel network of a back-of-the-book index

**Touria Aït El Mekki and Adeline Nazarenko**
LIPN —University of Paris13 & CNRS UMR 7030
Av. J.B. Clément, F-93430 Villetneuse, France
{taem,nazarenko}@lipn.univ-paris13.fr

## Abstract

Relying on the idea that back-of-the-book indexes are traditional devices for navigation through large documents, we have developed a method to build a hypertextual network that helps the navigation in a document. Building such an hypertextual network requires selecting a list of descriptors, identifying the relevant text segments to associate with each descriptor and finally ranking the descriptors and reference segments by relevance order. We propose a specific document segmentation method and a relevance measure for information ranking. The algorithms are tested on 4 corpora (of different types and domains) without human intervention or any semantic knowledge.

## 1 Introduction

Helping readers to get access to the document content is a text-mining challenge. Back-of-the-book indexes are traditional devices that provide an overview of the document content and help the reader to navigate through the document. An index[1] is "an alphabetical list of persons, places, subjects, etc., mentioned in the text of a printed work, usually at the back, and indicating where in the work they are referred to"[2]. More formally, an index is made of a nomenclature, which is a (structured) list of descriptors, and of a large set of references that link the descriptors to document segments. Such indexes are also designed for electronic documents and for web sites[3].

We have designed a method for automating the building of indexes. Our IndDoc system relies on the text of the document 1) to select the descriptors that are worth mentioning in the final index and 2) to link each descriptor to document segments. We do not address the first point here[4]. We rather focus on the elaboration of the hypertextual network.

Building such a network raises two problems. The first one is the *segmentation problem*. For each relevant descriptor, it is necessary to identify the relevant document segments to refer to. The difficult point is not to identify the various text occurrences of a descriptor, but to determine, for a given occurrence of a descriptor, to which span of text (short paragraph or whole section) it is necessary to refer. There is also a *relevance-ranking problem*. Linking all descriptors to all their occurrences would introduce too many links and work against navigation. A relevance measure must be defined to select the most important links.

Section 2 presents the previous works on navigational tools and segmentation or ranking methods. Our method is described in section 3. The section 4 presents our experiments and results.

## 2 Previous works

### 2.1 Existing indexing tools

Existing computer-aided indexing tools are either embedded in word processing or stand-alone software such as Macrex[5] and Cindex[4]. They are designed to assist a human indexer. They locate the various occurrences of a descriptor, automatically compute the page numbers for references, rank the entries in alphabetic order and format the resulting index according to a given index style sheet. However, the indexer still has to choose the relevant descriptors. In the best case, the indexing tool proposes a huge list of all the noun phrases to the indexer (*e.g.* Indexing online[5], Syntactica[6]). The indexer also has to identify the various forms under which a given descriptor is mentioned in the document and to select the descriptor occurrences that are worth referring to.

### 2.2 Navigation through a document

Various approaches have been developed to help readers to visualise large document bases (Byrd, 1999) but these methods are usually designed to handle IR results, *i.e.* rather large and potentially heterogeneous set of documents.

Less attention has been paid to the problem of navigating through a single document, which requires a finer grained content description due to the relative document homogeneity. Some document and collection browsers rely the list of the key phrases extracted from documents (Anick 01, Wacholder 01) but these works do not consider the document side of the index hypertextual network. (Gross & Assadi 97) presents a navigation system for a technical document but the method relies on a pre-existing ontology of the document domain. The indicative summaries (Saggion & Lapalme 02), which present the list of the keywords occurring in the most relevant phrases of the document, are close to traditional indexes but coarser-grained.

---

[1] In the following, the term *index* is always used with the same meaning.

[2] Collins 1998 dictionary définition.

[3] A web site can be considered as a special type of document and indexed in the same way as traditional printed books.

[4] It is based on a terminological analysis and includes the recognition of variant descriptors (Nazarenko & Aït El Mekki 2005).

[5] http://www.macrex.cix.co.uk/

[4] http://www.macrex.cix.co.uk/

[5] http://www.indexingonline.com/index.php

[6] http://www.syntactica.com/login/login1.htm

Independently of indexes, however, the segmentation and relevance-ranking problems are traditional ones in NLP and IR.

### 2.3 Segmentation approaches

Segmentation methods are usually based on the physical structure of the documents (typography, sectionning), on the lexical cohesion (Morris & Hirst 91; Hearst 97, Ferret *et al*. 98) and/or the linguistic markers expressing local continuity (Litman & Passonneau 95). The lexical cohesion approach gives interesting results on large and heterogeneous documents, but is less adapted to the segmentation of homogeneous documents. The structural and linguistic approaches are more relevant for our purposes. Our segmentation algorithm combines both methods (see Section 3).

However, traditional segmentation algorithm propose an absolute segmentation of documents, whereas, in indexes, the segmentation may vary from one entry to another. A whole set of paragraphs can be considered as a coherent Documentary Unit (UD) for a given entry and a smaller fragment be more relevant for another one.

### 2.4 Relevance measures

Ranking a set of documents is a well-known problem in IR. We adapted the traditional IR relevance tf.idf score (Salton 89) to rank the various paragraphs of a document instead of a set of documents.

The relevance problem is also addressed for document summarisation, to extract the more relevant sentences from the original document. The relevance score is based on the word weights, document structure and linguistic or typographical emphasis markers. Our relevance measure takes those parameters into account.

## 3 Method

For each descriptor, it is necessary to identify the relevant segments of the document that are worth referring to. This implies to detect its occurrences (not addressed here), identify the span of the segments to be referred to and to rank the results in relevance order.

### 3.1 Identifying reference segments

#### 3.1.1 Segmentation cues

Our segmentation method relies on the presence of markers of integration of structural, linguistic and typographical kind. The algorithm takes the following cues into account:
- The physical structure of texts (sectioning);
- The presence of markers of linear integration (*if, then, secondly, on the other han*d*, thus, moreover, in addition...*) at the beginning of a paragraph; IndDoc relies on a core dictionary of generic markers, which can be tuned and extended for any specific corpus;
- The presence of an anaphoric pronoun at the beginning of a paragraph: *this, this, these, it, its;*
- The lexical cohesion of contiguous paragraphs, which is based on the recurrence of the index descriptors and their variant and thesaurus relations for a fine-grained segmentation as opposed to (Hearst 97);
- The typographical homogeneity between contiguous paragraphs (two paragraphs in italics or several items of the same list, for instance).

#### 3.1.2 Segmentation algorithm

Our algorithm (Figure 1) is made up of two phases, which correspond to an absolute segmentation in *documentary units* (DU) and a relative segmentation in *reference segments*.

The *absolute segmentation* phase only depends on the document. We start with a rough segmentation of the document in minimal DUs (MDU) (step 1). These MDUs are then widened in DUs (step 2) according to the linguistic and typographical markers and to the logical structure of the document (a DU cannot cross a section frontier for instance). At the end of this phase, the document is represented as a list of DUs.

The *relative segmentation* phase depends on a given descriptor. It comprises three more steps. The segments of reference are first identified (DUs which contain an occurrence of the descriptor or of one of its variants) (step 3). The segments that are contiguous in the text of the document are then merged (step 4), which results in a simplified list of segments. The segments belonging to a same section are finally generalised into in a single reference to the whole section (step 5), if a significant part of the section is represented in the list of the segments established in step 2.

### 3.2 Relevance ranking

Our relevance measure is based on the tf.idf score. We apply it to the paragraphs of a text rather than to the documents of a given collection. We also adapted the tf.idf score to take into account, in addition to the weight of a word in the whole document and its frequency in the segment, the weight of a particular occurrence (which can be typographically emphasised, for example).

Two scores are taken into account: the descriptor score (d-score(i) for the descriptor $d_i$) and the segment score (s-score(i,j) for the the $j^{th}$ occurrence segment of $d_i$). A segment score is higher if it contains some important descriptors and a descriptor score is higher if it is mentioned in informative part of the document. We solve this traditional authority circularity problem by distinguishing in the following an intrinsic segment weight and a relative segment score.

---

Let $\mathcal{MDU}$ be the list of MDUs.

Let $\Sigma$ be the list of the all document sections and subsections.
Let $\mathcal{D} = \{d_1,...,d_m\}$ be the set of extracted descriptors.
Let $\mathcal{DU}$ be the list of DUs.
**Begin**
$\mathcal{DU} = \mathcal{MDU}$
               // Document Units
**For each** $du_i$ de $\mathcal{DU}$

Widen ud$_i$ to the next ud$_{i+1}$ of $\mathcal{DU}$
**if** there is no section frontier between ud$_i$ and ud$_{i+1}$
**and if** there is a linguistic or typographical continuity between ud$_i$ and ud$_{i+1}$.
// Plain segments
**For each** d$_i$ descriptor of $\mathcal{D}$:
Compute d$_i^+$, the class formed by d$_i$ and its variant forms.
**For each** d$_i^+$ class of $\mathcal{D}^+$:
Compute $\mathcal{S}_i^+$, the list of the DUs in which the d$_i^+$ descriptors occur.
// Simplified segments
Compute $\mathcal{SS}_i^+$ from $\mathcal{S}_i^+$ by merging the contiguous segments.
//Generalised segments
**For each** σ$_j$ of Σ
Identify the set e$_{ij}$ of all segments of $\mathcal{SS}_i^+$ belonging to σ$_j$.
**if** the proportion of occurrences of the d$_i^+$ descriptors *per* paragraph in the section σ$_j$ is higher than a given threshold,
**then** the section σ$_j$ as a whole is considered as a reference segment for d$_j^+$ and the e$_{ij}$ paragraph sublist is substituted by σ$_j$ in Σ.
**else** each paragraph of e$_{ij}$ is considered as an individual reference segment for d$_j^+$.
**End.**
The **linguistic continuity** is marked by the presence of a marker listed in the dictionary of linear integration
The **typography continuity** is marked by italic, bold or list structure

Figure 1: The segmentation algorithm

### 3.2.1 Segment score

The *s-score(i,j)* is defined by the following formula:

$$s-score(i,j) = siw_j \cdot \sum_{k=1}^{D}(\alpha \cdot sdw(i,j))$$

where D is the total number of descriptors in the document and α = 1 if d$_k$ is d$_i$ or one of its variants and 0,5 otherwise.

The score of the segment $s_{ij}$, *s-score(i,j)* is based on two elementary weights. (1) The *segment informational weight* (*siw$_j$*) is intrinsic to the segment $s_j$. It is high if $s_j$ contains some typographical markers (bold, italics…) or new descriptors (first occurrence in $s_j$). It also depends on the status of the segment in the document: titles are more relevant segments than the summary or the conclusion. (2) The *segment discriminating weight* of the segment $s_j$ relatively to the descriptor $d_i$ (*sdw$_{ij}$*) depends on the number of occurrences of $d_i$ in $s_j$ and of its distribution over the document. *ssw$_{ij}$* is high if $d_i$ has several occurrences in $s_j$ and if it mainly occurs in $s_j$. This weight is a revised tf.idf measure:

$$sdw_{ij} = occ_{ij} \cdot \log(p/p_i)$$

where $occ_{ij}$ is the number of occurrences of $d_i$ in $s_j$, P is the total number of paragraphs in the document and $P_i$ is the number of paragraphs in which $d_i$ occurs.

### 3.2.2 Descriptor score

The d-score(i) is defined by the following formula:

$$d-score(i) = dsw_i \cdot \sqrt{ddw_i \cdot diw_i \cdot \sum_{j=1}^{p_i} s-score(i,j)/p_i}$$

The score of the descriptor $d_i$, *d-score(i)* is based on three elementary weights. (1) The *descriptor informational weight* (*diw$_i$*) depends on the typographical characteristics of individual occurrences of $d_i$ and of the weights of the segments in which it occurs. *diw$_i$* is high if some occurrences of $d_i$ are typographically emphasised or if $d_i$ appears in special document parts (such as the titles, summary, introduction…). (2) The *descriptor discriminating weight* (*ddw$_i$*) depends on the normalised number of occurrences $d_i$ and of its distribution over the document. *dsw$_i$* is high is $d_i$ occurs more often than the other descriptors and if it is irregularly distributed. This weight is a revised tf.idf measure.

$$ddw_i = {occ_i}/{occ'} \cdot \log(p/p_i)$$

where *occ'* is the mean number of occurrences per descriptor. (3) The *descriptor semantic weight* (*dsw$_i$*) depends on the number of descriptors to which $d_i$ is linked in the semantic network of the index nomenclature.

Relevance is thus computed from a large set of cues. Besides frequency, typography, document structure, distribution and semantic network density are exploited.

## 4 Experiments and results

### 4.1 Corpora

Our first experiments are based on four different French corpora (Table 1): 2 handbooks in artificial intelligence (AI) and linguistics (LI) and 2 collections of scientific papers dealing with Knowledge Engineering (in the following: KE01 and KE04).

|  | Monographs | | Collections | |
|---|---|---|---|---|
|  | **LI** | **AI** | **KE01** | **KE04** |
| Corpus size (# words occurrences) | 42 260 | 111 371 | 185 382 | 122 229 |
| Vocabulary size (without empty words) | 3 018 | 9 429 | 38 962 | 32 334 |
| Nomenclature size (# descriptors) | 615 | 1 361 | 10 008 | 8 259 |
| Corpus size (# paragraphs) | 793 | 7 386 | 4 929 | 5 110 |

Table 1: Corpus profiles

|  | Unit types | Unit number | | | |
|---|---|---|---|---|---|
|  |  | KE04 | KE01 | AI | LI |
| 1 | Min. Doc. Units | 5110 | 4929 | 7386 | 793 |
| 2 | Doc. Units | 4272 | 4698 | 7245 | 634 |
| 3 | Plain segments | 14585 | 9863 | 8823 | 2569 |
| 4 | Simplified segments | 13876 | 9786 | 5157 | 1893 |
| 5 | Generalised segments | 13345 | 9728 | 4469 | 950 |
| 6 | Paragraph occurrences | 39089 | 18974 | 9897 | 3983 |

| | Reduction factors | | | |
|---|---|---|---|---|
|  | KE04 | KE01 | AI | LI |
| 1->2 | -20% | -10% | -0% | 30% |
| 3->4 | -10% | -0% | -40% | -30% |
| 4->5 | -10% | -10% | -20% | -50% |
| 5->6 | -33% | -50% | -45% | -25% |

Table 2: Segmentation results

## 4.2 Segmentation

### 4.2.1 Example

The Figure 2 presents a segmentation example. The initial text is divided into 4 paragraphs (4 MDUs). Because of the presence of markers of linear integration (*Actually, Moreover*), the MDU corresponding to the paragraph §i is widened to cover *§i-§i+2*. The absolute segmentation thus gives 2 DUs : *§i-§i+2* and *§i+3*. For the relative segmentation, let us consider the descriptor "contexte d'insertion" (*insertion context*). The only occurrence of that descriptor in the whole document appears in paragraph *§i* (DU *§i-§i+2*). This single reference segment is finally generalised to the whole section because the segment of reference covers three of the four paragraphs of the section.

---

section k : Begin

§i     Le **contexte d'insertion** d'une ACCA a nécessairement des incidences ….

§i+1    **En effet** (Actually), pour atteindre …

§i+2    **De plus** (Moreover), même si dans notre cas le domaine est une variable libre, il faut qu'il ….

§i+3    Ces différentes considérations nous ont conduit à proposer une activité,….

section k : End

---

Figure 2: A segmentation exemple

### 4.2.2 Global segmentation behaviour

We applied the segmentation algorithm to our four corpora. The results are given in Table 2. The left part of the table describes the lists of textual units obtained at each step. The segmentation reduces the number of references for each corpus. The 6$^{th}$ line (size of the corpus in terms of paragraph number) is added for comparison: we consider the number of paragraphs as a basic segmentation reference. The comparison between the lines 5 and 6 shows that our segmentation algorithm actually reduces the number of references (from 25% to 50%) but we observe that:

- The reduction factors (right part of the table) depend on the nature of the document (monograph *vs* collection) and of their style;
- The simplification of segments (line 3->4) has a stronger effect on monographs due to lexical homogeneity;
- For the KE corpora, which are rather heterogeneous, the first step (line 1->2) is the more important.
- There are proportionally more integration markers in LI than in AI.
- The segment generalisation has a stronger impact on LI, which is more strictly structured in sections and subsections.

The diversity of the segmentation cues make our segmentation algorithm robust to various types of documents

## 4.3 Relevance ranking

Our relevance ranking algorithm behaves as expected on our experimental corpora .

### 4.3.1 Example

Let us consider the descriptor "contrainte temporelle" (*temporal constraint*). The 12 initial occurrences of this descriptor in LI corpus are grouped into 3 reference segments during the segmentation phase:

- S1 contains the first occurrence of the descriptor which is written in bold and which is a definition but it is a small segment.
- S2 is composed of three subsections. "contrainte temporelle" occurs in the title of the first one and is mentioned in the two others. The descriptors "concordance des temps" (*sequence of tenses*) and "relation temporelle" (*temporal relation*") which are semantically close[9] to "contrainte temporelle" occur in the titles of the second and third subsections.
- The descriptor appears at the beginning of the third segment but S3 itself belongs to a conclusion.

The ranking gives the references in the following order: S2, S1 and S3. S2 is given first because it is the most informative and it contains a title occurrence of the descriptor. Even if S1 contains the first occurrence of the descriptor and if it is typographically emphasized, it is considered as less informative. The segment S3 is last because it is a conclusion part.

It is interesting to consider the "contrainte temporelle" entry in the index of the published LI book. The published index gives exactly the same segments (along with an empty and probably erroneous reference), in textual order, which is less informative.

### 4.3.2 Segment ranking evaluation

To evaluate our segment ranking measure, we have selected a sample of 30 descriptors that have numerous reference segments among the 110 descriptors of the original published LI index. For each descriptor, the author of the book was asked to analyse the quality of the segment ranking.

The results are given in Table 3. We distinguished the descriptors whose segment list is correctly ranked (group 1), those for which the ranking is only partially correct (but the top list is good, group 2), those whose ranking is globally incorrect (group 3) and the undecidable cases (group 4).

Table 3 shows that the top of the segment lists are correct in 77% of the cases and that the ranking algorithm fails in less than 15% of the cases. A detailed analysis shows that defining occurrences tend to get high-ranking scores: for polysemous descriptors (such as *origine* (Engl. *origin*)), the technical occurrences are better ranked than the common sense ones (*à l'origine de/to begin with*).

---

[9] These semantic relations are computed during the terminological analysis that is note presented here (Nazarenko & Aït El Mekki 05).

| Correct ranking : 17% | | Incorrect ranking: 23% | |
|---|---|---|---|
| Group 1 | Group 2 | Goup 3 | Group 4 |
| 17 | 6 | 4 | 3 |

Table 3: Segment ranking for 30 descriptors

### 4.3.3 Descriptor ranking evaluation

The ranking of the descriptors does not have direct impact on navigation functionalities but the ranking of segments and descriptors are interdependent.

For evaluation purposes, an independent indexer was asked to choose the most relevant descriptors in the flat list of 615 LI descriptors. She decided to keep 203 descriptors. If we consider the ranking of those 203 relevant descriptors, we observe that the mean rank is 126,5, which is much higher than the 307,5 median rank. The precision at the 203rd position in the ranking is 83%. For the KE04 experiment, only the 1500 top ranked descriptors have been validated and the precision rate is 70%. For test purposes, 500 descriptors with low scores have been artificially added. All but one of these "bad" descriptors were actually eliminated (less than 0.01% of precision).

Those figures confirm the rather good performance of our knowledge-poor ranking algorithm.

## 5 Conclusion

We propose a knowledge poor method to automatically build the hypertextual network that helps the navigation through the document. The resulting device is similar to an back-of-the-book index. We show that, given a document and a list of descriptors, it is possible to automatically compute a network of reference links that connect the list of descriptors to the text of the document. Two interrelated problems must be solved: What are the spans of text that are worth referring to for each descriptor? What are the most relevant pieces of information (descriptors and references) for navigation? We adapted the traditional techniques developed for text segmentation and document ranking. The originality of our method is the large variety of cues that are taken into account: typography, document logical structure, linguistic markers of linear integration, lexical cohesion, etc. The impact of each type of cue depends on the document style but the combination of all make our segmentation and ranking algorithm more robust.